\documentclass[11pt]{article}
\usepackage[margin=1in]{geometry}
\usepackage{amsmath,amssymb}
\usepackage{graphicx}
\usepackage[round]{natbib}
\usepackage{url}
\usepackage{longtable}
\usepackage{booktabs}
\usepackage{wrapfig}
\usepackage{worldflags}
\usepackage[load-configurations=version-1]{siunitx}
\usepackage{tikz}
\usetikzlibrary{arrows,matrix,positioning}
\usetikzlibrary{fit}
\usetikzlibrary{arrows.meta,arrows}
\usetikzlibrary{decorations.pathmorphing}
\usetikzlibrary{calc}
\usepackage{pgfplots}
\pgfplotsset{compat=1.17}
\usepackage[colorlinks=true,linkcolor=blue,citecolor=blue,urlcolor=blue]{hyperref}
\usepackage{cleveref}
\graphicspath{{./figures/}}

\title{LLM-Guided Graph Generation for\\ Structure-Based Local Improvement Methods}
\author{Hai Xia\qquad Vaidyanathan Peruvemba Ramaswamy\qquad Stefan Szeider\\[4pt]
  Algorithms and Complexity Group, TU Wien, Austria\\
  \texttt{\{hxia,vaidyanathan,sz\}@ac.tuwien.ac.at}}
\date{}

\begin{document}

\maketitle

\begin{abstract}
  Large neighborhood search normally selects a random subset of decision variables for iterative optimization. To efficiently solve various problems, researchers tend to design variable selection strategies that take into account structural features across different domains. In this paper, we build an automatic pipeline that is problem-agnostic to all problems in the MiniZinc format. By prompting an LLM with our semantic guidelines, we guide the LLM to produce a graph generator that maps any instance of a problem type to a uniform weighted graph, where nodes represent decision variables and edges represent constraint relationships. These problem-agnostic graphs guide our structure-based local improvement (SLIM) framework for variable selection. Meanwhile, the weighted graph enables all problem instances to share the same generic graph representation, from which the same graph features can be extracted and used for configuration selection. We evaluated our pipeline on instances across 20 MiniZinc competition problems, finding that algorithm selection achieves a 39.6\% average problem-weighted win rate against a one-shot Gurobi baseline, more than doubling the best single configuration (19.3\%). A post-hoc configuration and a feature ablation indicate a headroom of up to 44.0\%, demonstrating that LLM-based semantic generation enables effective automated structure and feature extraction for constraint optimization.
\end{abstract}

\section{Introduction}

Large Neighborhood Search (LNS) is one of the most useful
metaheuristics for hard combinatorial optimization
problems~\citep{Shaw1998,PisingerRopke2019}, especially when the
problems become larger and more complicated.
In each iteration, LNS selects only a subset of the decision variables,
which is optimized by an exact solver.
After the local instance is optimized, the better local solution is
patched back to the original global one.
Therefore, the efficiency of the search relies heavily on whether
promising and suitable variables are selected for local optimization.

In practice, variable selection (neighborhood selection) is quite
flexible.
Random selection is the simplest one, which is computationally cheap,
but ignores the instance features and the problem structure, wasting
solving time on loosely coupled subproblems.
Structure-guided variable selection can exploit the variable relations,
producing tightly coupled neighborhoods.
But what kind of features or structure information should be considered
is also a question quite related to the specific problem type, and it
needs a lot of expert domain knowledge.
For different problem domains, there are different principles for
designing efficient Structure-Based Local Improvement Methods
(SLIM)~\citep{FichteLodha2017}.
In the last decade, SLIM has been applied to different problem domains,
including treewidth computation~\citep{FichteLodha2017},
branchwidth~\citep{Lodha2019}, treedepth~\citep{Vaidyanathan2020},
Bayesian network learning~\citep{Vaidyanathan2021,Vaidyanathan2022},
graph coloring~\citep{SchidlerSzeider2023}, decision tree
optimization~\citep{SchidlerSzeider2024jair}, and Maximum
Satisfiability~\citep{SchidlerSzeider2024cp}.
In different problem-specific SLIM algorithms, there are various
construction methods for neighborhoods and corresponding variable
selection strategies.
All these designs are highly engineered by domain experts, taking
months of manual work.
Besides the algorithm design, efficient algorithm implementation is
also not trivial when applied to different problem instances: algorithm
configurators typically demand extensive computational budgets for
getting better configurations of SLIM
algorithms~\citep{Vaidyanathan2024}, and portfolio-based algorithm
selection requires problem-specific features that also need to be
curated by experts~\citep{xia24sat}.

To alleviate the difficulty of problem-specific application of SLIM, we
build a pipeline where we can easily build structure-aware SLIM
algorithms for different problems with the help of a Large Language
Model (LLM).
First, by giving the MiniZinc\footnote{MiniZinc is a solver-independent constraint modeling language: \url{https://www.minizinc.org}} constraint models to the LLM, we can
prompt with our semantic guidelines to guide the LLM to produce a
Python program (a \emph{graph generator}) that can map all
problem-specific instances to a uniform weighted generic graph.
In the generic graph, the nodes correspond to decision variables and
take weights and domain sizes.
Meanwhile, the edges represent the constraint relations between different
nodes, carrying coupling strengths.
Therefore, the graph has no problem-specific properties, and it is a
problem-agnostic structure representation.
Then SLIM can operate on the uniform graph by some basic extraction
algorithms (extraction based on breadth-first search (BFS) and weighted random sampling).
The generic extraction algorithms can select the neighborhoods with
semantic structure information derived from the original constraints.
Furthermore, the problem-generic representation also makes it possible
to do cross-problem algorithm selection, as we can extract topological
and statistical features from the generic graphs of the instances from
different problems.

In brief, we have the contributions as follows.
\begin{enumerate}
  \item We propose using an LLM to produce validated graph
    generators covering different problems. With the graph
    generator, generic graphs can be generated in a uniform way for
    instances from different problems.
  \item By using the graph generators, we build problem-agnostic SLIM
    algorithms easily across different problems without specific expert
    knowledge.
  \item We introduce cross-problem configuration selection using
    generic graph-based features. From the extensive evaluation on 20
    MiniZinc competition problems, we find
    that our model achieves a 39.6\% average
    problem-weighted win rate against a one-shot Gurobi baseline, more
    than doubling the best single configuration (19.3\%), with a
    post-hoc ablation analysis indicating potential of up to 44.0\%.
\end{enumerate}

\section{Related Work}

\paragraph{Large neighborhood search.}
\citet{Shaw1998} introduced LNS for vehicle routing, showing that structure-aware
destruction outperforms random selection. \citet{PisingerRopke2019} survey the area,
including adaptive LNS~\citep{RopkePisinger2006}. Recent neural variants learn the
destroy/repair policy~\citep{HottungTierney2022,Johnn2023} but remain tied to specific
problem types. In contrast, we generate problem-agnostic graph structures offline and
apply them to any MiniZinc problem type after a one-time generator synthesis, without
any per-problem neural training.

\paragraph{Structure-guided local improvement.}
We build on SLIM, introduced for treewidth by \citet{FichteLodha2017} and since extended
to branchwidth~\citep{Lodha2019}, treedepth~\citep{Vaidyanathan2020}, Bayesian network
learning~\citep{Vaidyanathan2021,Vaidyanathan2022}, graph coloring~\citep{SchidlerSzeider2023},
decision trees~\citep{SchidlerSzeider2024jair}, and Maximum
Satisfiability~\citep{SchidlerSzeider2024cp}. Each application required substantial
domain-specific engineering, but we automate this step with an LLM, turning SLIM into a
problem-agnostic methodology without needing too much domain knowledge.

\paragraph{Structure from constraint models, and LLMs for optimization.}
Exploiting constraint-graph topology for solving is classical~\citep{DechterPearl1989,Gottlob2002},
and recent work learns branching policies from variable--constraint
graphs~\citep{Gasse2019}. These extract \emph{syntactic} structure, whereas our LLM assigns
weights reflecting semantic roles. LLMs have been used to discover
programs~\citep{RomeraParedes2024}, as iterative optimizers~\citep{Yang2024}, to generate
constraint models from natural language~\citep{Text2Zinc2025}, and to enforce constraints
during LLM decoding~\citep{Bonlarron2025}. We instead use the LLM as a
one-time offline compiler that produces deterministic, auditable generators.

\paragraph{Algorithm selection.}
Portfolio-based selection is well studied~\citep{Xu2008,Lindauer2015}, typically
within a single domain using domain-specific features. Because all our instances share one
uniform graph representation, a single model selects configurations across 20 heterogeneous
problem types without per-problem feature engineering. The supplementary
material gives an extended discussion.

\section{Methodology}
\label{sec:methodology}

\subsection{Pipeline Overview}
\label{sec:pipeline}

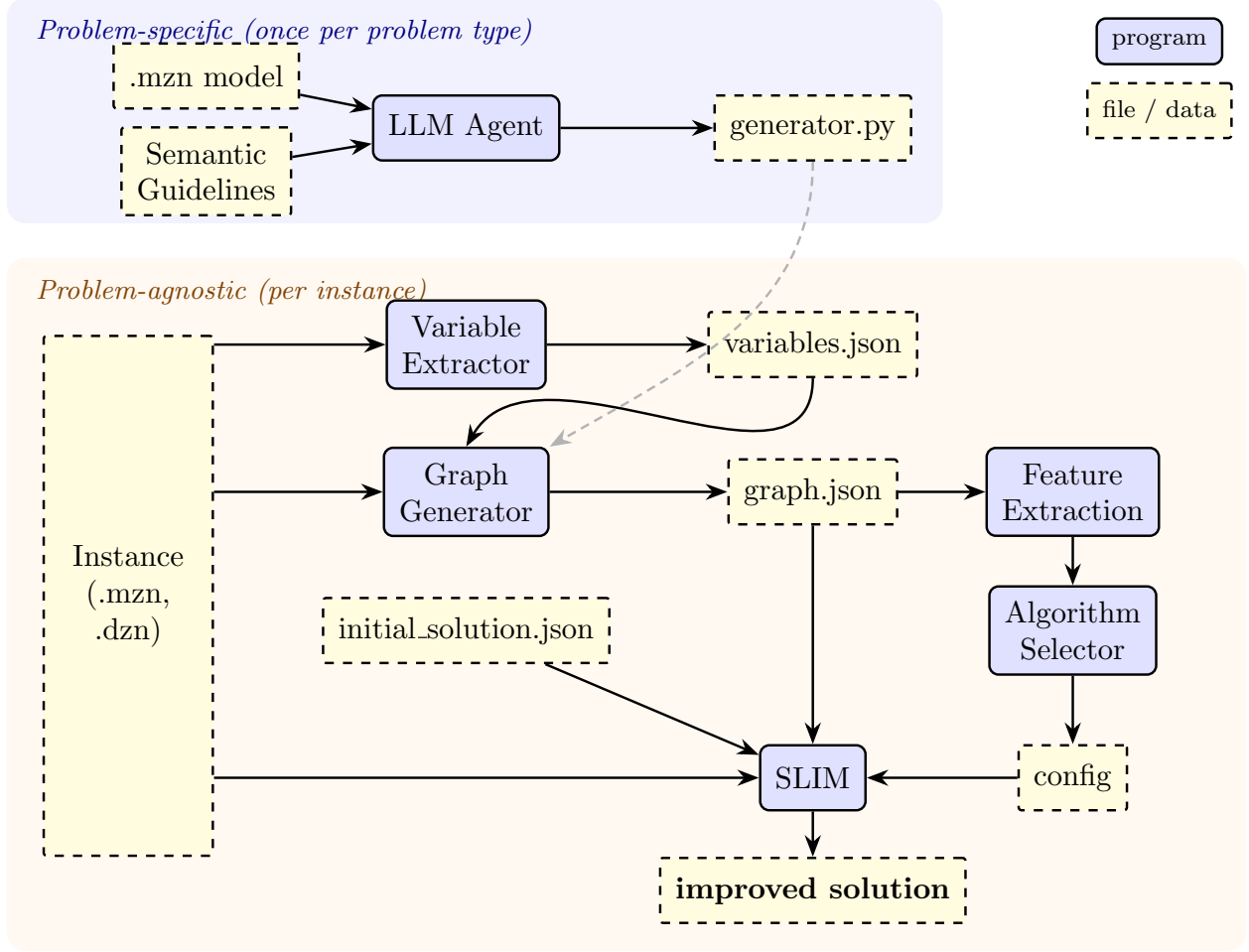
\begin{figure}[t]
  \centering
  \resizebox{\textwidth}{!}{%
  \begin{tikzpicture}[
    % Programs/modules: solid thick border, colored fill
    prog/.style={draw, thick, rounded corners=3pt, fill=blue!12,
                 minimum height=0.75cm, font=\small, inner sep=5pt, align=center},
    % Files/data: dashed border, cream fill
    file/.style={draw, thick, dashed, rounded corners=1pt, fill=yellow!15,
                 minimum height=0.75cm, font=\small, inner sep=5pt, align=center},
    % Arrows
    arr/.style={-{Stealth[length=2.5mm]}, thick},
    darr/.style={-{Stealth[length=2.5mm]}, thick, densely dashed, gray!60},
    >=Stealth
  ]
    % === BACKGROUND REGIONS ===
    \fill[blue!5, rounded corners=6pt] (-0.8, 4.5) rectangle (10, 1.9);
    \fill[orange!5, rounded corners=6pt] (-0.8, 1.5) rectangle (13.5, -6.5);

    % Phase labels
    \node[font=\footnotesize\itshape, blue!50!black, anchor=north west] at (-0.6, 4.4)
      {Problem-specific (once per problem type)};
    \node[font=\footnotesize\itshape, orange!50!black, anchor=north west] at (-0.6, 1.4)
      {Problem-agnostic (per instance)};

    % === PROBLEM-SPECIFIC PHASE ===
    % .mzn + Guidelines (stacked left) → LLM (center) → generator.py (right)
    \node[file] (mzn) at (1.5, 3.6) {.mzn model};
    \node[file] (guide) at (1.5, 2.5) {Semantic\\Guidelines};
    \node[prog] (llm) at (4.5, 3.0) {LLM Agent};
    \node[file] (genpy) at (8.5, 3.0) {generator.py};

    \draw[arr] (mzn) -- (llm);
    \draw[arr] (guide) -- (llm);
    \draw[arr] (llm) -- (genpy);

    % === PROBLEM-AGNOSTIC PHASE ===

    % Instance files: tall box on the left spanning rows
    \node[file, minimum height=6cm, text width=1.6cm, align=center] (inst) at (0.6, -2.4)
      {Instance\\(.mzn,\\.dzn)};

    % Row 1: Variable extraction
    \node[prog] (varext) at (4.5, 0.5) {Variable\\Extractor};
    \node[file] (varjson) at (8.5, 0.5) {variables.json};
    \draw[arr] (inst.east |- varext) -- (varext);
    \draw[arr] (varext) -- (varjson);

    % Row 2: Graph generation
    \node[prog] (gen) at (4.5, -1.2) {Graph\\Generator};
    \node[file] (graph) at (8.5, -1.2) {graph.json};
    \draw[arr] (inst.east |- gen) -- (gen);
    \draw[arr] (gen) -- (graph);
    % variables.json feeds generator
    \draw[arr] (varjson.south) to[out=-90, in=60] (gen.north);
    % generator.py instantiates the Generator
    \draw[darr] (genpy.south) to[out=-90, in=30] (gen.north east);

    % Row 3: Feature extraction and algorithm selection (right column)
    \node[prog] (feat) at (11.5, -1.2) {Feature\\Extraction};
    \node[prog] (algsel) at (11.5, -2.8) {Algorithm\\Selector};
    \node[file] (config) at (11.5, -4.5) {config};
    \draw[arr] (graph) -- (feat);
    \draw[arr] (feat) -- (algsel);
    \draw[arr] (algsel) -- (config);

    % Row 4: SLIM
    \node[file] (init) at (4.5, -2.8) {initial\_solution.json};
    \node[prog] (slim) at (8.5, -4.5) {SLIM};
    \node[file, font=\small\bfseries] (improved) at (8.5, -5.8) {improved solution};
    \draw[arr] (inst.east |- slim) -- (slim);
    \draw[arr] (graph.south) to[out=-90, in=90] (slim.north);
    \draw[arr] (init) -- (slim);
    \draw[arr] (config.west|-slim) -- (slim.east);
    \draw[arr] (slim) -- (improved);

    % Legend (upper right)
    \node[prog, minimum height=0.45cm, font=\scriptsize] (lprog) at (12.5, 4.0) {program};
    \node[file, minimum height=0.45cm, font=\scriptsize] (lfile) at (12.5, 3.2) {file / data};
  \end{tikzpicture}%
  }
  \caption{Pipeline overview. Programs (solid border) process files (dashed border). The blue region shows the one-time generator creation by the LLM, and the orange region shows per-instance processing. The instance files (\texttt{.mzn} + \texttt{.dzn}) feed the Variable Extractor, the Graph Generator, and SLIM (which solves subproblems on the original model). The dashed arrow indicates that \texttt{generator.py} is instantiated as the Graph Generator.}
  \label{fig:pipeline}
\end{figure}

For each MiniZinc constraint model, we can generate a Python program (a
\emph{graph generator}) with the help of the LLM, which can be used for
transforming different instances of the problem into a generic graph representation.
Therefore, the generic graph can be used for guiding the variable
selection in the SLIM framework and enabling algorithm configuration
selection in a problem-agnostic fashion according to the uniform graph features.
To this end, the whole pipeline has two parts: the
\emph{problem-specific} part and the \emph{problem-agnostic} part.
\Cref{fig:pipeline} illustrates the general mechanism of our framework.

In the problem-specific stage, the LLM (Claude Opus 4.5\footnote{\url{https://www.anthropic.com/claude}}) reads the MiniZinc model (\texttt{.mzn})
for a given problem type together with our semantic weight guidelines
and produces a \emph{graph generator}, a Python program
that understands the semantic structure of that problem's
constraints and variables.
This is a one-time effort for each problem: we synthesize generators
for all problems, and they are applicable to all instances from
MiniZinc competitions (2008--2025).
Of these, instances across 20 problem types satisfy our benchmarking criteria (see \Cref{sec:exp-eval}).
Each synthesis for a single problem usually takes within several
minutes, freeing experts from curating their own problem-specific SLIM
algorithms.

In the problem-agnostic stage, the variable extractor parses the
decision variables with their domains and array indices from the
MiniZinc instances and the model, producing a \texttt{variables.json}
file.
With the parsed \texttt{variables.json} file, the LLM-generated graph
generator can construct weighted graphs (with the instance
information), wherein nodes represent decision variables (with
importance weights and domain sizes) and edges indicate the constraint
relations (with coupling strengths).
Therefore, SLIM can also use the generic graph for selecting variables
according to the weighted neighborhoods for local improvements.
For example, the BFS-based and the random-based extraction algorithms
can operate on the generic graph, which is from a vehicle routing problem (VRP) instance or a
scheduling instance.
The problem domain information is implicitly contained in the generic
graph.
And the generic problem-agnostic SLIM algorithm has no idea what domain
problem it is solving.
Based on the same generic graph representation, the graph features,
like the topological and statistical features of the uniform graph, are
also extracted and then used for algorithm configuration selection
(among 30 SLIM configurations) for better solving efficiency.

\subsection{LLM-Guided Graph Generation}
\label{sec:graph-generation}

The general idea for the graph generation is using the LLM as a
\emph{semantic compiler}.
With the guidelines set by human researchers, the LLM can generate a
Python program (the graph generator), where the instance can be mapped
into a weighted graph.
The semantic approach can capture the relationships between the
variables and constraints.
For example, we can set a general range in the prompts regarding the
emphasis of different constraints: the variables in an
\texttt{alldifferent} constraint over 50 elements should be connected
with light edges to avoid clique domination.

\subsubsection{Problem-Agnostic Graph Format}
For each instance, the graph generator can produce a uniform graph
representation:
\begin{itemize}
  \item \textbf{Nodes} indicate decision variables. Each node has two
    properties: a weight $w \in [0,1]$ (importance to the objective)
    and a domain size $d \in \mathbb{N}$ (number of possible values).
  \item \textbf{Edges} indicate constraint relations between different
    decision variables, each with a weight $w \in [0,1]$ showing how
    the decision variables are coupled.
\end{itemize}
The uniform representation allows different components of SLIM and
algorithm selection to work in a generic way.
For example, the budget computation is counted according to the number
of nodes selected and the corresponding domain sizes.

\subsubsection{Generation Guidelines}

Here are our defined guidelines for the LLM to generate targeted graph
generators with semantic information:
\begin{enumerate}
  \item \textbf{Objective-related components should have higher
    weights} ($\geq 0.6$), because they usually have direct influence
    on the final objective values. So these components should have
    higher weights to be selected by the extraction algorithms during
    the SLIM optimization phase.
  \item \textbf{Lower weights for large global constraints}: as there
    are some global constraints among the decision variables, if we
    give high weights to all these global constraints, then the
    weights of different components will be similar, resulting in no
    preference during the variable selection. Large constraints
    (e.g., \texttt{alldifferent} over $n > 10$ variables) use weight
    $\max(0.1,\; 1/n)$ to prevent clique domination, while small
    constraints ($n \leq 5$) retain strong coupling ($\geq 0.8$).
  \item \textbf{No isolated nodes}, as the graph-based extraction can
    only reach the nodes that are connected by different edges. We
    have to make sure all nodes have the possibility to be selected
    and optimized without any search space left out.
  \item \textbf{Bounded weights}: to make the extraction algorithms
    operate in a uniform way, we set an upper bound for the weights.
    When there are several constraints linking the same variables,
    edge weights are aggregated according to
    $W = 1 - \prod_i (1 - w_i)$, resulting in weights never
    exceeding~1.
\end{enumerate}

A worked example of the graph-generation process is given in the supplementary material.

\subsection{Generic Structure-Based Local Improvement (SLIM)}
\label{sec:slim}

As we already have a generic graph representation from the graph
generator, SLIM can now work on a problem-agnostic uniform weighted
graph.
Even though SLIM now has no prior knowledge or preferences for
different problems, it can extract the variables with some useful
information, like the weights and the domain sizes of the nodes, and
the coupling weights of the edges in the graph.
In each iteration, we use the extraction algorithms to select the
neighborhood of variables from the uniform graph, while the remaining
variables outside of the neighborhood are frozen.
Then the local solver can optimize the smaller subproblem, resulting in
better local solutions.

\subsubsection{Extraction Methods}
Our SLIM has two extraction methods based on the generic graph:
\begin{itemize}
  \item \textbf{BFS extraction} starts from a randomly chosen node
    according to the node weights, and then expands nodes via edges
    in a weight-biased random order. These can capture the locality, with
    considerations of both the importance of the decision variables
    and the constraint structures of the original problem.
  \item \textbf{LNS extraction} collects random variables using the
    corresponding node weights, but it has no consideration regarding
    the locality preferences like BFS does, where a node can be
    expanded only when it is connected with the current collected
    component by an edge.
\end{itemize}
Both of the extractions stop when the current budget is reached: the
\emph{budget} parameter~$b$ is a threshold bounded by a
domain-size-aware metric.
Besides the collection of variables to be optimized, we also have
another freeze mode, which is for collecting the variables to be
frozen instead of to be optimized. We have a detailed description in
the supplementary material.

In the whole SLIM working loop (see the supplementary material for the pseudocode), it starts from an
initial solution~$s_0$, which is usually a suboptimal solution
obtained by heuristics or by running the solver with a short timeout.
Then SLIM extracts a neighborhood~$N$ from the graph using the
extraction method $\mathcal{E}$ (BFS or LNS) with the given
budget~$b$.
Meanwhile, the freeze mode~$f$ determines whether the variables
collected should be frozen or should be optimized.
To get a better solution, the bounding constraint obj$(s') \leq$
obj$(s^*)$ (for minimization; $\geq$ for maximization) is added to the
subproblem.
Therefore, after the per-iteration timeout~$t$, if the solver returns
a feasible solution that is not worse than the current incumbent, the
global solution can be updated.
SLIM accepts equal-quality solutions, and this allows us to explore
different solutions with the same objective value.
With this working mechanism, the loop continues until the overall time
budget is exhausted or the maximum number of iterations is reached.

In the SLIM framework, there are several configuration parameters
having huge influence on the final performance.
This is also one of the reasons why we have the following algorithm
configuration selection for boosting the performance.
For example, the local budget and the local time indicate different
preferences for different local structures to be optimized.
In our setting, we evaluate 30 configurations in total, including the
local budget $b \in \{10, 20, 50, 70, 100, 200\}$, the timeout
$t \in \{20, 30, 45, 60\}$\,s, and different extraction strategies.

\subsection{Algorithm Selection}
\label{sec:algorithm-selection}

As we discussed, the 30 SLIM configurations with different strategies
have various preferences for solving problem instances.
Note that even though there are sophisticated algorithms
selectors~\citep{Xu2008,Lindauer2015}, they are usually
applicable to instances of the same problem. There are also some related works where
MiniZinc models are flattened to get a uniform representation, from which uniform features can be extracted~\citep{Amadini2014}.
In our setting, we train a machine learning model to select the
best configuration for different instances across problems, as we have
a uniform graph representation, from which we can extract the
graph features from the original variables.
We built five simple selection approaches to demonstrate the
effectiveness of our pipeline. Any advances in algorithm selection
would further improve our results.

There are different methods for training the algorithm selector. Here we only use the regression approach as a representative example to show how the pipeline works (see the supplementary material for the pseudocode).
In the training phase, the feature vector~$\mathbf{x}_i$ is first extracted
from each instance's graph, and per-configuration improvement
margins~$\mathbf{y}_i$ are the differences between each
configuration's improvement and the one-shot baseline
improvement.
Then we train a multi-output random forest regression model with problem-weighted samples as different problems have quite different numbers of instances. After the cross-validation, the final model is retrained on the full training set.
Next, we could evaluate how good the portfolio algorithm is by applying the algorithm predicted by the machine learning model.
The other four algorithm selectors proceed with the same workflow, and the only difference is about the training loss objective and what they actually predict during the test. The supplementary material lists the summary of the five different algorithm selectors.

In our feature extraction setting, we design a feature set consisting of 54 features computed from the weighted uniform graph of each instance together with lightweight instance metadata (see the supplementary material). The feature set includes topology statistics,
node and edge weight distributions,
domain size statistics and correlations, variable metadata, and numeric instance parameters.
As we have the uniform graph format across problems, the same 54 features work identically for all problem instances. Note that the dataset is heavily imbalanced (the resource-constrained project scheduling problem, RCPSP, alone accounts for 38\%).
To prevent specific problems from dominating the evaluation, we assign problem-weighted sample weights so that each of the problems contributes equally during training as well as the final evaluation on the test set.

In the ablation analysis on the feature set (54 features) and the algorithm portfolio set (30 configurations), we apply greedy backward elimination. The corresponding experimental results are in the supplementary material. The configuration ablation removes configurations whose elimination improves the final performance on the test instances,
and feature ablation subsequently removes features from the reduced configuration set. We report this as a post-hoc analysis of the pruning potential.

\section{Experimental Evaluation}
\label{sec:exp-eval}

In this section, we show the experimental analysis of our pipeline on different problems.
\subsection{Experimental Settings}

As we mentioned in \Cref{sec:pipeline}, we collect problems from MiniZinc competitions\footnote{\url{https://www.minizinc.org/challenge/}} (2008--2025) resulting in instances across 20 problem types after we filter out unsuitable instances. We have several criteria for the selection: 1. We include instances where Gurobi (a commercial exact solver) finds a feasible solution within 10 minutes but does not prove optimality within 60 minutes. 2. We only include problems with at least 5 qualifying instances.
For algorithm selection, we split the instances with a 70:30 ratio, stratified by problem type, resulting in the instance distribution on training and test sets as shown in
the supplementary material.

We run all experiments on a Sun Grid Engine cluster with 20 nodes running Ubuntu 18.04 LTS.
Each node has two Intel Xeon E5-2640 v4 2.40\,GHz CPUs and 160\,GB RAM. SLIM and related programs use Python 3.6.9.
The Zenodo repository\footnote{\url{https://doi.org/10.5281/zenodo.21910103}} contains materials for reproducing the results.

\subsection{SLIM vs One-Shot Gurobi}

We compare our problem-agnostic SLIM against the one-shot Gurobi baseline. The one-shot Gurobi means running Gurobi on the original MiniZinc
instance with the same total time budget (60 minutes). As
Gurobi is a strong industrial commercial solver, it has been widely used on many problems. We compare the solutions generated by Gurobi's default running and the solutions optimized with our problem-agnostic SLIM.
As we introduced in \Cref{sec:slim}, there are different strategies that can be used during running. Therefore, SLIM using Gurobi as the subsolver becomes standard LNS when the variable extraction is random and does not utilize the uniform weights.

\begin{figure}[h]
  \centering
  \includegraphics[width=\textwidth]{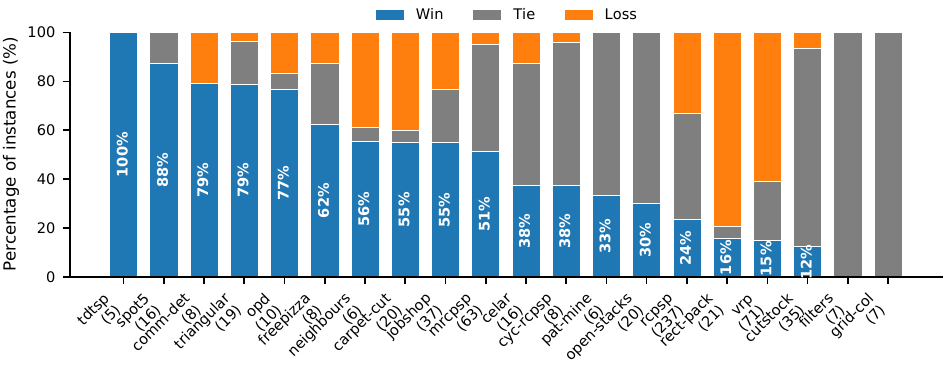}
  \caption{SLIM (best configuration per problem type) vs one-shot Gurobi baseline, averaged across 3 seeds. Problem types are sorted by win rate. Instance counts per type are shown in parentheses.}
  \label{fig:slim-vs-oneshot}
\end{figure}

We show detailed comparison results among the 20 problems in
\Cref{fig:slim-vs-oneshot}.
On each problem, we use SLIM to run on all instances with the 30
configurations, and then we can compare the objective values of SLIM
with the best-performing configuration against those generated by
one-shot Gurobi.
From the win/tie/loss percentages on each problem, we can see that on
some problems, problem-agnostic SLIM can get high-performing results.
On problems including tdtsp, spot5, community-detection, triangular,
and opd, SLIM can win over 75\% of the instances.
Among all 20 problems, there are only two problems on which SLIM gets
worse solutions on more than 50\% of the instances: rectangle-packing
and VRP.

Even though SLIM failed to get any better solutions on the filters and
grid-coloring problems, on all instances of these two problems SLIM
can get the same good solutions as one-shot Gurobi.
In general, when SLIM is equipped with suitable strategies, it can get substantially better solutions than the one-shot Gurobi. Meanwhile, the results also indicate that the configurations (suitable search strategies) are important, which
motivates our problem-agnostic algorithm selection idea in the following section. We could use algorithm selection to learn a good mapping from different instances to the configurations for guiding the structure-guided search.

\subsection{Algorithm Selection}

We use the aforementioned five algorithm selection approaches across three random seeds (51, 52, 53). In each training run of the algorithm selectors, we use the same train/test instance distribution for a fair comparison.

In \Cref{tab:approaches-comparison}, we show the problem-weighted win rates of all five selection approaches against the best single configuration baseline.
All five approaches substantially outperform the baseline across all seeds: the best single configuration reaches only 17.8--22.3\% depending on the seed, while every selection approach exceeds 31\%.
Regarding the performance across seeds, no single approach dominates
others.
The two algorithm selection approaches, regression and ensemble, win
on two of the three seeds.
It indicates that per-configuration improvement margins are more
suitable for training the algorithm configuration selectors of the
problem-agnostic SLIM.
Even though the binary algorithm selection approach is the
best-performing selector on only one seed, it is the most stable
algorithm selector, whose win rate fluctuates within two percentage
points.
We can see that for all algorithm selection approaches, the win rate
is always below 50\%.
Actually, it does not mean weak performance among the problems, as
they still have a high tie rate for each problem.
The detailed win/tie/loss rates are also shown in the supplementary
material. In general, for the best-performing approach on each seed, the problem-weighted win rates are tightly clustered (37.9--40.6\%) with positive net scores (win minus loss of 14.8--24.5\%). And the performance of the portfolio algorithm is well above the best single configuration, and reasonably below the virtual best.

\begin{table}[h]
  \centering
  \caption{Problem-weighted win rate (\%) of all selection approaches across 3 seeds.}
  \label{tab:approaches-comparison}
  \small
  \begin{tabular}{lccc}
    \toprule
    Approach & Seed 51 & Seed 52 & Seed 53 \\
    \midrule
    regression     & \textbf{40.1} & 40.2 & 34.2 \\
    ensemble       & 38.3 & \textbf{40.6} & 35.3 \\
    binary         & 36.4 & 36.9 & \textbf{37.9} \\
    classification & 36.9 & 39.0 & 36.2 \\
    two\_stage     & 34.3 & 37.2 & 31.9 \\
    \midrule
    Best single config & 17.8 & 17.7 & 22.3 \\
    \bottomrule
  \end{tabular}
\end{table}

% \subsection{Discussion}
% We discuss our choice of one-shot Gurobi as a baseline, the cost and practicality of generator creation, and threats to validity in Appendix~\ref{app:discussion}.

\section{Conclusion}

In this paper, we presented an automatic pipeline where an LLM is used
for synthesizing graph generators to transform MiniZinc problem
instances into a uniform graph representation.
With the problem-generic graph, we can use structure-aware SLIM across
different problems.
Meanwhile, we can also extract the features from the uniform graph to
automatically select the best-performing configurations for SLIM.
By extensive evaluation on the 20 problems, we found that the
problem-agnostic SLIM can get stronger results compared with one-shot
Gurobi.

There are also limitations.
When the problem-agnostic SLIM is applied to specific problems, SLIM
is not able to outperform one-shot Gurobi even with the best search
strategies.
This is probably due to the LLM-guided graph generation. The generators, while validated, remain LLM-produced approximations of the true constraint semantics, and our evaluation covers only MiniZinc competition benchmarks. In the future, we may have more targeted prompts to guide the LLM to
transform the instances into uniform graphs with other types of
information, like some probing topological features and so on.
Furthermore, we can even add some information that can be dynamically
updated during the search.
For example, after specific components have been optimized many times,
we can use tabu search or adaptively adjust the weights to escape the
repetitive optimization.

\subsection*{Acknowledgements}

\leavevmode
\begin{wrapfigure}[3]{l}{25pt}{\worldflag[width=18pt]{EU}}
\end{wrapfigure}
\noindent This project is also partially supported by the European Union's Horizon 2020 research and innovation programme
under the Marie Skłodowska-Curie grant agreement No.~101034440, and by the Austrian Science Fund (FWF) within the Cluster of Excellence Bilateral Artificial Intelligence (10.55776/COE12) and 10.55776/P36420.

\bibliographystyle{plainnat}
\bibliography{literature}

\begin{thebibliography}{25}
\providecommand{\natexlab}[1]{#1}
\providecommand{\url}[1]{\texttt{#1}}
\expandafter\ifx\csname urlstyle\endcsname\relax
  \providecommand{\doi}[1]{doi: #1}\else
  \providecommand{\doi}{doi: \begingroup \urlstyle{rm}\Url}\fi

\bibitem[Amadini et~al.(2014)Amadini, Gabbrielli, and Mauro]{Amadini2014}
Roberto Amadini, Maurizio Gabbrielli, and Jacopo Mauro.
\newblock {SUNNY:} a lazy portfolio approach for constraint solving.
\newblock \emph{Theory Pract. Log. Program.}, 14\penalty0 (4-5):\penalty0
  509--524, 2014.
\newblock \doi{10.1017/S1471068414000179}.
\newblock URL \url{https://doi.org/10.1017/S1471068414000179}.

\bibitem[Bonlarron et~al.(2025)Bonlarron, R{\'{e}}gin, Maria, and
  R{\'{e}}gin]{Bonlarron2025}
Alexandre Bonlarron, Florian R{\'{e}}gin, Elisabetta~De Maria, and
  Jean{-}Charles R{\'{e}}gin.
\newblock Large language model meets constraint propagation.
\newblock In \emph{Proceedings of the Thirty-Fourth International Joint
  Conference on Artificial Intelligence, {IJCAI} 2025, Montreal, Canada, August
  16-22, 2025}, pages 10036--10044. ijcai.org, 2025.
\newblock \doi{10.24963/IJCAI.2025/1115}.
\newblock URL \url{https://doi.org/10.24963/ijcai.2025/1115}.

\bibitem[Dechter and Pearl(1989)]{DechterPearl1989}
Rina Dechter and Judea Pearl.
\newblock Tree clustering for constraint networks.
\newblock \emph{Artif. Intell.}, 38\penalty0 (3):\penalty0 353--366, 1989.
\newblock \doi{10.1016/0004-3702(89)90037-4}.
\newblock URL \url{https://doi.org/10.1016/0004-3702(89)90037-4}.

\bibitem[Fichte et~al.(2017)Fichte, Lodha, and Szeider]{FichteLodha2017}
Johannes~Klaus Fichte, Neha Lodha, and Stefan Szeider.
\newblock {SAT}-based local improvement for finding tree decompositions of
  small width.
\newblock In Serge Gaspers and Toby Walsh, editors, \emph{Theory and
  Applications of Satisfiability Testing - {SAT} 2017 - 20th International
  Conference, Melbourne, VIC, Australia, August 28 - September 1, 2017,
  Proceedings}, volume 10491 of \emph{Lecture Notes in Computer Science}, pages
  401--411. Springer, 2017.
\newblock \doi{10.1007/978-3-319-66263-3\_25}.
\newblock URL \url{https://doi.org/10.1007/978-3-319-66263-3\_25}.

\bibitem[Gasse et~al.(2019)Gasse, Ch{\'{e}}telat, Ferroni, Charlin, and
  Lodi]{Gasse2019}
Maxime Gasse, Didier Ch{\'{e}}telat, Nicola Ferroni, Laurent Charlin, and
  Andrea Lodi.
\newblock Exact combinatorial optimization with graph convolutional neural
  networks.
\newblock In Hanna~M. Wallach, Hugo Larochelle, Alina Beygelzimer, Florence
  d'Alch{\'{e}}{-}Buc, Emily~B. Fox, and Roman Garnett, editors, \emph{Advances
  in Neural Information Processing Systems 32: Annual Conference on Neural
  Information Processing Systems 2019, NeurIPS 2019, December 8-14, 2019,
  Vancouver, BC, Canada}, pages 15554--15566, 2019.
\newblock URL
  \url{https://proceedings.neurips.cc/paper/2019/hash/d14c2267d848abeb81fd590f371d39bd-Abstract.html}.

\bibitem[Gottlob et~al.(2002)Gottlob, Leone, and Scarcello]{Gottlob2002}
Georg Gottlob, Nicola Leone, and Francesco Scarcello.
\newblock Hypertree decompositions and tractable queries.
\newblock \emph{J. Comput. Syst. Sci.}, 64\penalty0 (3):\penalty0 579--627,
  2002.
\newblock \doi{10.1006/JCSS.2001.1809}.
\newblock URL \url{https://doi.org/10.1006/jcss.2001.1809}.

\bibitem[Hottung and Tierney(2022)]{HottungTierney2022}
Andr{\'{e}} Hottung and Kevin Tierney.
\newblock Neural large neighborhood search for routing problems.
\newblock \emph{Artif. Intell.}, 313:\penalty0 103786, 2022.
\newblock \doi{10.1016/J.ARTINT.2022.103786}.
\newblock URL \url{https://doi.org/10.1016/j.artint.2022.103786}.

\bibitem[Johnn et~al.(2023)Johnn, Darvariu, Handl, and Kalcsics]{Johnn2023}
Syu{-}Ning Johnn, Victor{-}Alexandru Darvariu, Julia Handl, and J{\"{o}}rg
  Kalcsics.
\newblock {GRAPH} reinforcement learning for operator selection in the {ALNS}
  metaheuristic.
\newblock In Bernab{\'{e}} Dorronsoro, Francisco Chicano, Gr{\'{e}}goire Danoy,
  and El{-}Ghazali Talbi, editors, \emph{Optimization and Learning - 6th
  International Conference, {OLA} 2023, Malaga, Spain, May 3-5, 2023,
  Proceedings}, volume 1824 of \emph{Communications in Computer and Information
  Science}, pages 200--212. Springer, 2023.
\newblock \doi{10.1007/978-3-031-34020-8\_15}.
\newblock URL \url{https://doi.org/10.1007/978-3-031-34020-8\_15}.

\bibitem[Lindauer et~al.(2015)Lindauer, Hoos, Hutter, and Schaub]{Lindauer2015}
Marius Lindauer, Holger~H. Hoos, Frank Hutter, and Torsten Schaub.
\newblock Autofolio: An automatically configured algorithm selector.
\newblock \emph{J. Artif. Intell. Res.}, 53:\penalty0 745--778, 2015.
\newblock \doi{10.1613/JAIR.4726}.
\newblock URL \url{https://doi.org/10.1613/jair.4726}.

\bibitem[Lodha et~al.(2019)Lodha, Ordyniak, and Szeider]{Lodha2019}
Neha Lodha, Sebastian Ordyniak, and Stefan Szeider.
\newblock A {SAT} approach to branchwidth.
\newblock \emph{{ACM} Trans. Comput. Log.}, 20\penalty0 (3):\penalty0
  15:1--15:24, 2019.
\newblock \doi{10.1145/3326159}.
\newblock URL \url{https://doi.org/10.1145/3326159}.

\bibitem[{Peruvemba Ramaswamy} and Szeider(2022)]{Vaidyanathan2022}
Vaidyanathan {Peruvemba Ramaswamy} and Stefan Szeider.
\newblock Learning large {Bayesian} networks with expert constraints.
\newblock In James Cussens and Kun Zhang, editors, \emph{Proceedings of the
  Thirty-Eighth Conference on Uncertainty in Artificial Intelligence, {UAI}
  2022}, volume 180 of \emph{Proceedings of Machine Learning Research}, pages
  1592--1601. {PMLR}, 2022.
\newblock URL
  \url{https://proceedings.mlr.press/v180/peruvemba-ramaswamy22a.html}.

\bibitem[Pisinger and Ropke(2019)]{PisingerRopke2019}
David Pisinger and Stefan Ropke.
\newblock Large neighborhood search.
\newblock In Michel Gendreau and Jean{-}Yves Potvin, editors, \emph{Handbook of
  Metaheuristics}, pages 99--127. Springer International Publishing, Cham,
  2019.
\newblock ISBN 978-3-319-91086-4.
\newblock \doi{10.1007/978-3-319-91086-4_4}.
\newblock URL \url{https://doi.org/10.1007/978-3-319-91086-4_4}.

\bibitem[Ramaswamy and Szeider(2020)]{Vaidyanathan2020}
Vaidyanathan~Peruvemba Ramaswamy and Stefan Szeider.
\newblock {MaxSAT-Based} postprocessing for treedepth.
\newblock In Helmut Simonis, editor, \emph{Principles and Practice of
  Constraint Programming - 26th International Conference, {CP} 2020,
  Louvain-la-Neuve, Belgium, September 7-11, 2020, Proceedings}, volume 12333
  of \emph{Lecture Notes in Computer Science}, pages 478--495. Springer, 2020.
\newblock \doi{10.1007/978-3-030-58475-7\_28}.
\newblock URL \url{https://doi.org/10.1007/978-3-030-58475-7\_28}.

\bibitem[Ramaswamy and Szeider(2021)]{Vaidyanathan2021}
Vaidyanathan~Peruvemba Ramaswamy and Stefan Szeider.
\newblock Turbocharging treewidth-bounded bayesian network structure learning.
\newblock In \emph{Thirty-Fifth {AAAI} Conference on Artificial Intelligence,
  {AAAI} 2021, Thirty-Third Conference on Innovative Applications of Artificial
  Intelligence, {IAAI} 2021, The Eleventh Symposium on Educational Advances in
  Artificial Intelligence, {EAAI} 2021, Virtual Event, February 2-9, 2021},
  pages 3895--3903. {AAAI} Press, 2021.
\newblock \doi{10.1609/AAAI.V35I5.16508}.
\newblock URL \url{https://doi.org/10.1609/aaai.v35i5.16508}.

\bibitem[Ramaswamy et~al.(2024)Ramaswamy, Szeider, and Xia]{Vaidyanathan2024}
Vaidyanathan~Peruvemba Ramaswamy, Stefan Szeider, and Hai Xia.
\newblock The power of collaboration: Learning large bayesian networks at
  scale.
\newblock In \emph{36th {IEEE} International Conference on Tools with
  Artificial Intelligence, {ICTAI} 2024, Herndon, VA, USA, October 28-30,
  2024}, pages 371--378. {IEEE}, 2024.
\newblock \doi{10.1109/ICTAI62512.2024.00061}.
\newblock URL \url{https://doi.org/10.1109/ICTAI62512.2024.00061}.

\bibitem[Romera{-}Paredes et~al.(2024)Romera{-}Paredes, Barekatain, Novikov,
  Balog, Kumar, Dupont, Ruiz, Ellenberg, Wang, Fawzi, Kohli, and
  Fawzi]{RomeraParedes2024}
Bernardino Romera{-}Paredes, Mohammadamin Barekatain, Alexander Novikov, Matej
  Balog, M.~Pawan Kumar, Emilien Dupont, Francisco J.~R. Ruiz, Jordan~S.
  Ellenberg, Pengming Wang, Omar Fawzi, Pushmeet Kohli, and Alhussein Fawzi.
\newblock Mathematical discoveries from program search with large language
  models.
\newblock \emph{Nature}, 625\penalty0 (7995):\penalty0 468--475, 2024.
\newblock \doi{10.1038/S41586-023-06924-6}.
\newblock URL \url{https://doi.org/10.1038/s41586-023-06924-6}.

\bibitem[Ropke and Pisinger(2006)]{RopkePisinger2006}
Stefan Ropke and David Pisinger.
\newblock An adaptive large neighborhood search heuristic for the pickup and
  delivery problem with time windows.
\newblock \emph{Transp. Sci.}, 40\penalty0 (4):\penalty0 455--472, 2006.
\newblock \doi{10.1287/TRSC.1050.0135}.
\newblock URL \url{https://doi.org/10.1287/trsc.1050.0135}.

\bibitem[Schidler and Szeider(2023)]{SchidlerSzeider2023}
Andr{\'{e}} Schidler and Stefan Szeider.
\newblock Sat-boosted tabu search for coloring massive graphs.
\newblock \emph{{ACM} J. Exp. Algorithmics}, 28:\penalty0 1.5:1--1.5:19, 2023.
\newblock \doi{10.1145/3603112}.
\newblock URL \url{https://doi.org/10.1145/3603112}.

\bibitem[Schidler and Szeider(2024{\natexlab{a}})]{SchidlerSzeider2024cp}
Andr{\'{e}} Schidler and Stefan Szeider.
\newblock Structure-guided local improvement for maximum satisfiability.
\newblock In Paul Shaw, editor, \emph{30th International Conference on
  Principles and Practice of Constraint Programming, {CP} 2024, Girona, Spain,
  September 2-6, 2024}, volume 307 of \emph{LIPIcs}, pages 26:1--26:23. Schloss
  Dagstuhl - Leibniz-Zentrum f{\"{u}}r Informatik, 2024{\natexlab{a}}.
\newblock \doi{10.4230/LIPICS.CP.2024.26}.
\newblock URL \url{https://doi.org/10.4230/LIPIcs.CP.2024.26}.

\bibitem[Schidler and Szeider(2024{\natexlab{b}})]{SchidlerSzeider2024jair}
Andr{\'{e}} Schidler and Stefan Szeider.
\newblock {SAT}-based decision tree learning for large data sets.
\newblock \emph{J. Artif. Intell. Res.}, 80:\penalty0 875--918,
  2024{\natexlab{b}}.
\newblock \doi{10.1613/JAIR.1.15956}.
\newblock URL \url{https://doi.org/10.1613/jair.1.15956}.

\bibitem[Shaw(1998)]{Shaw1998}
Paul Shaw.
\newblock Using constraint programming and local search methods to solve
  vehicle routing problems.
\newblock In Michael~J. Maher and Jean{-}Francois Puget, editors,
  \emph{Principles and Practice of Constraint Programming - CP98, 4th
  International Conference, Pisa, Italy, October 26-30, 1998, Proceedings},
  volume 1520 of \emph{Lecture Notes in Computer Science}, pages 417--431.
  Springer, 1998.
\newblock \doi{10.1007/3-540-49481-2\_30}.
\newblock URL \url{https://doi.org/10.1007/3-540-49481-2\_30}.

\bibitem[Singirikonda et~al.(2025)Singirikonda, Kadioglu, and
  Uppuluri]{Text2Zinc2025}
Akash Singirikonda, Serdar Kadioglu, and Karthik Uppuluri.
\newblock Text2zinc: {A} cross-domain dataset for modeling optimization and
  satisfaction problems in minizinc.
\newblock \emph{CoRR}, abs/2503.10642, 2025.
\newblock \doi{10.48550/ARXIV.2503.10642}.
\newblock URL \url{https://doi.org/10.48550/arXiv.2503.10642}.

\bibitem[Xia and Szeider(2024)]{xia24sat}
Hai Xia and Stefan Szeider.
\newblock {SAT-Based} tree decomposition with iterative cascading policy
  selection.
\newblock In Michael~J. Wooldridge, Jennifer~G. Dy, and Sriraam Natarajan,
  editors, \emph{Thirty-Eighth {AAAI} Conference on Artificial Intelligence,
  {AAAI} 2024, Thirty-Sixth Conference on Innovative Applications of Artificial
  Intelligence, {IAAI} 2024, Fourteenth Symposium on Educational Advances in
  Artificial Intelligence, {EAAI} 2024, February 20-27, 2024, Vancouver,
  Canada}, pages 8191--8199. {AAAI} Press, 2024.
\newblock \doi{10.1609/AAAI.V38I8.28659}.
\newblock URL \url{https://doi.org/10.1609/aaai.v38i8.28659}.

\bibitem[Xu et~al.(2008)Xu, Hutter, Hoos, and Leyton{-}Brown]{Xu2008}
Lin Xu, Frank Hutter, Holger~H. Hoos, and Kevin Leyton{-}Brown.
\newblock Satzilla: Portfolio-based algorithm selection for {SAT}.
\newblock \emph{J. Artif. Intell. Res.}, 32:\penalty0 565--606, 2008.
\newblock \doi{10.1613/JAIR.2490}.
\newblock URL \url{https://doi.org/10.1613/jair.2490}.

\bibitem[Yang et~al.(2024)Yang, Wang, Lu, Liu, Le, Zhou, and Chen]{Yang2024}
Chengrun Yang, Xuezhi Wang, Yifeng Lu, Hanxiao Liu, Quoc~V. Le, Denny Zhou, and
  Xinyun Chen.
\newblock Large language models as optimizers.
\newblock In \emph{The Twelfth International Conference on Learning
  Representations, {ICLR} 2024, Vienna, Austria, May 7-11, 2024}.
  OpenReview.net, 2024.
\newblock URL \url{https://openreview.net/forum?id=Bb4VGOWELI}.

\end{thebibliography}

% \newpage
% \begin{center}
%   {\LARGE\bfseries Supplementary Material}
% \end{center}
% \vspace{0.5em}

\end{document}